\documentclass[letterpaper]{article} 
\usepackage{aaai2026}  
\usepackage{times}  
\usepackage{helvet}  
\usepackage{courier}  
\usepackage[hyphens]{url}  
\usepackage{graphicx} 
\usepackage{natbib}  
\usepackage{caption} 
\usepackage{algorithm}
\usepackage{algorithmic}

\usepackage{amsfonts}

\usepackage{amsmath}
\usepackage{booktabs}

\usepackage{newfloat}
\usepackage{listings}
\DeclareCaptionStyle{ruled}{labelfont=normalfont,labelsep=colon,strut=off} 
\floatstyle{ruled}
\newfloat{listing}{tb}{lst}{}
\floatname{listing}{Listing}
\title{SLCFormer: Spectral-Local Context Transformer with Physics-Grounded Flare Synthesis for Nighttime Flare Removal}
\author{
    Xiyu Zhu\textsuperscript{\rm 1,2},
    Wei Wang\textsuperscript{\rm 1,2}\thanks{Corresponding author: wangwei8@wust.edu.cn},
    Xin Yuan\textsuperscript{\rm 2},
    Xiao Wang\textsuperscript{\rm 1}
}
\affiliations{

    \textsuperscript{\rm 1}School of Computer Science and Technology, Wuhan University of Science and Technology

    \textsuperscript{\rm 2}Hubei Province Key Laboratory of Intelligent Information Processing and Real-time
    
    Industrial System, Wuhan University of Science and Technology

    xiyu\_zhu@wust.edu.cn, wangwei8@wust.edu.cn, xinyuan@wust.edu.cn, wangxiao2021@wust.edu.cn
}

\usepackage{bibentry}

\begin{document}

\maketitle

\begin{abstract}
Lens flare is a common nighttime artifact caused by strong light sources scattering within camera lenses, leading to hazy streaks, halos, and glare that degrade visual quality. However, existing methods usually fail to effectively address nonuniform scattered flares, which severely reduces their applicability to complex real-world scenarios with diverse lighting conditions.
To address this issue, we propose SLCFormer, a novel spectral-local context transformer framework for effective nighttime lens flare removal. SLCFormer integrates two key modules: the Frequency Fourier and Excitation Module (FFEM), which captures efficient global contextual representations in the frequency domain to model flare characteristics, and the Directionally-Enhanced Spatial Module (DESM) for local structural enhancement and directional features in the spatial domain for precise flare removal. Furthermore, we introduce a ZernikeVAE-based scatter flare generation pipeline to synthesize physically realistic scatter flares with spatially varying PSFs, bridging optical physics and data-driven training. Extensive experiments on the Flare7K++ dataset demonstrate that our method achieves state-of-the-art performance, outperforming existing approaches in both quantitative metrics and perceptual visual quality, and generalizing robustly to real nighttime scenes with complex flare artifacts.
\end{abstract}

\section{Introduction}
Lens flare is a common optical phenomenon that is generated primarily from strong light entering a camera lens and scattering through several internal components such as lens surfaces, coatings, or iris mechanisms \cite{ernst2005interactive, hullin2011physically}. These artifacts are especially prominent at night, where bright light sources like streetlights or headlights contrast sharply with dark backgrounds, severely degrading image quality and hindering downstream vision tasks \cite{shao2025iebins, li2025neural} such as object detection \cite{2014rcnn}, autonomous driving \cite{2016autodriving} and depth estimation.

\begin{figure}[t] 
\centering
\includegraphics[width=\linewidth]{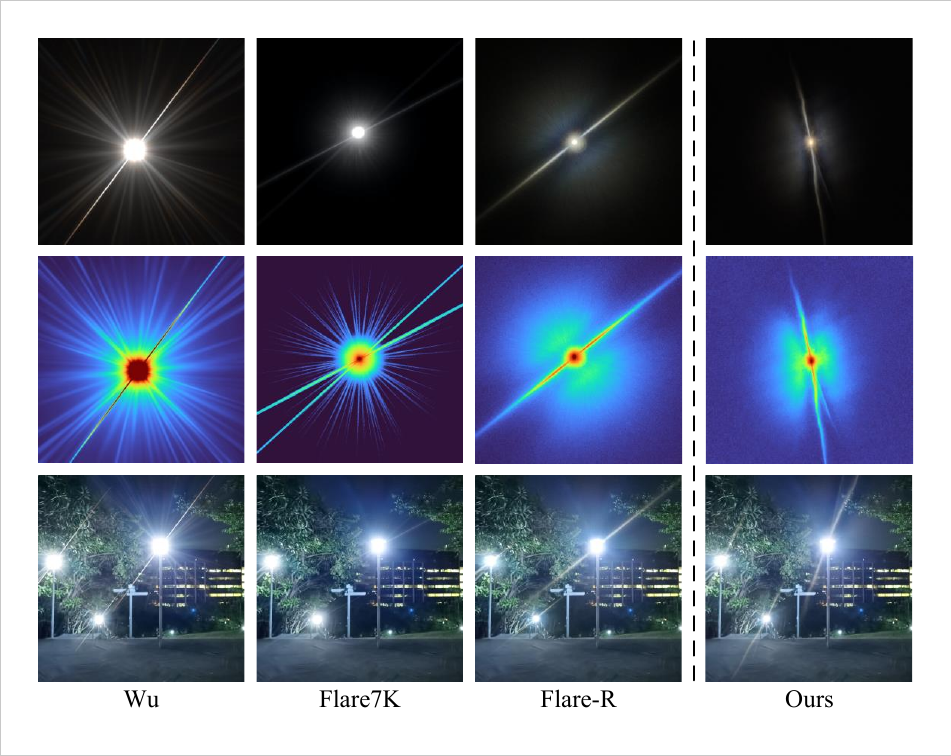} 
\caption{The comparison of scattering flare. We improve on the original dataset by embedding optical aberrations in the PSF for spatially varying scattering flares, as visualized in the intensity distribution heatmap.}
\label{fig:right_top}
\end{figure}

To address the lens flare problem, researchers have used traditional hardware and software solutions. Hardware designs modify optical systems (lens coatings or aperture adjustments) to suppress flare \cite{hardsoftlenschange}, but they are costly and inflexible. Software-based techniques rely on brightness thresholding \cite{vitoria2019automatic}, edge cues \cite{EdgeDetection}, or image decomposition, yet they only handle simple flare patterns and struggle with diverse real-world distributions.

With the rise of deep learning in computer vision \cite{Attentionisallyouneed,zamir2022restormer}, flare removal techniques have been driven from experiential rules to end-to-end learning frameworks. An important benchmark in this transition is the work of \cite{HowToTrain} who debuted the first large-scale synthetic dataset for lens flare removal, enabling supervised learning across diverse flare types and lighting conditions. They used a semi-synthetic approach to synthesize flare images with clean scenes, which greatly improved the feasibility of deep learning-based methods for flare removal. Building on this foundation, the Flare7K \cite{Flare7k} and enhanced Flare7K++ \cite{Flare7Kpp} datasets incorporate more different flare categories (including glare, light sources, and reflection artifacts), enriching the diversity and realism of the training data. 

Recent methods have explored increasingly sophisticated architectures to address the multifaceted nature of flare artifacts. Despite some progress in recent years, most existing flare removal models still face significant problems in representational and practical applicability. Many methods treat flare removal as a traditional image restoration problem without explicitly distinguishing flare artifacts from scene content. Additionally, many networks use attention-based modules \cite{liu2021swin}, which are computationally expensive, though they are powerful. Some recent attempts have employed multi-branch designs to preserve better detail, but the cost of architecture and training instability increased. These challenges highlight the need for more efficient flare removal frameworks that are adaptable to a wide range of flare types while preserving image fidelity and remaining computationally tractable.

To solve this problem, we first construct a ZernikeVAE-based scatter flare generation pipeline for enhancing the complexity of the dataset with the realism of synthetic flares (shown in Fig. \ref{fig:right_top}). The method generates the diffracted flare patterns by modeling the spatially varying point spread function (PSF) to break the limitation of the simple superposition of traditional synthetic flare. Unlike previous works that assume radially symmetric or uniform flare distributions, this is the first to capture non-uniform flares lacking central symmetry in both intensity and direction.
On this basis, we propose SLCFormer, a hybrid frequency-space transformer for nighttime flare removal. It integrates two core modules: FFEM, which leverages Fourier transform and channel attention to efficiently capture global frequency features of flare, and DESM, which enhances local structural and directional modeling via directionally-aware convolutions. In summary, our contributions are as follows:
\begin{itemize}
    \item A ZernikeVAE-based scatter flare generation pipeline that synthesizes more realistic and complex scatter flares by modeling physically spatially varying PSFs.
    
    \item We design SLCFormer, a novel spectral-local context transformer architecture that integrates frequency-domain global context modeling with spatial-domain directional feature enhancement for effective flare removal.
    
    \item Our method achieves superior performance compared to most existing models on both synthetic and real-world datasets.
\end{itemize}

\section{Related Work}
\subsection{Datasets for Flare Removal}
Earlier studies tended to use synthetic datasets, which limit the generalizability of the models. \cite{HowToTrain} proposed a semi-synthetic dataset, but the diversity and realism of the captured flares were limited because they were acquired under the same conditions. To address this, the Flare7K \cite{Flare7k} dataset introduced a more complex flare pattern using optical flares that simulate real-world scattering and reflective flares. Dai's follow-up result, Flare7K++ \cite{Flare7Kpp}, integrated real flares from contaminated lenses and enhanced the dataset's ability to model both scattered and reflected flares, resulting in better training for flare removal.

\begin{figure}[t] 
\centering
\includegraphics[width=\linewidth]{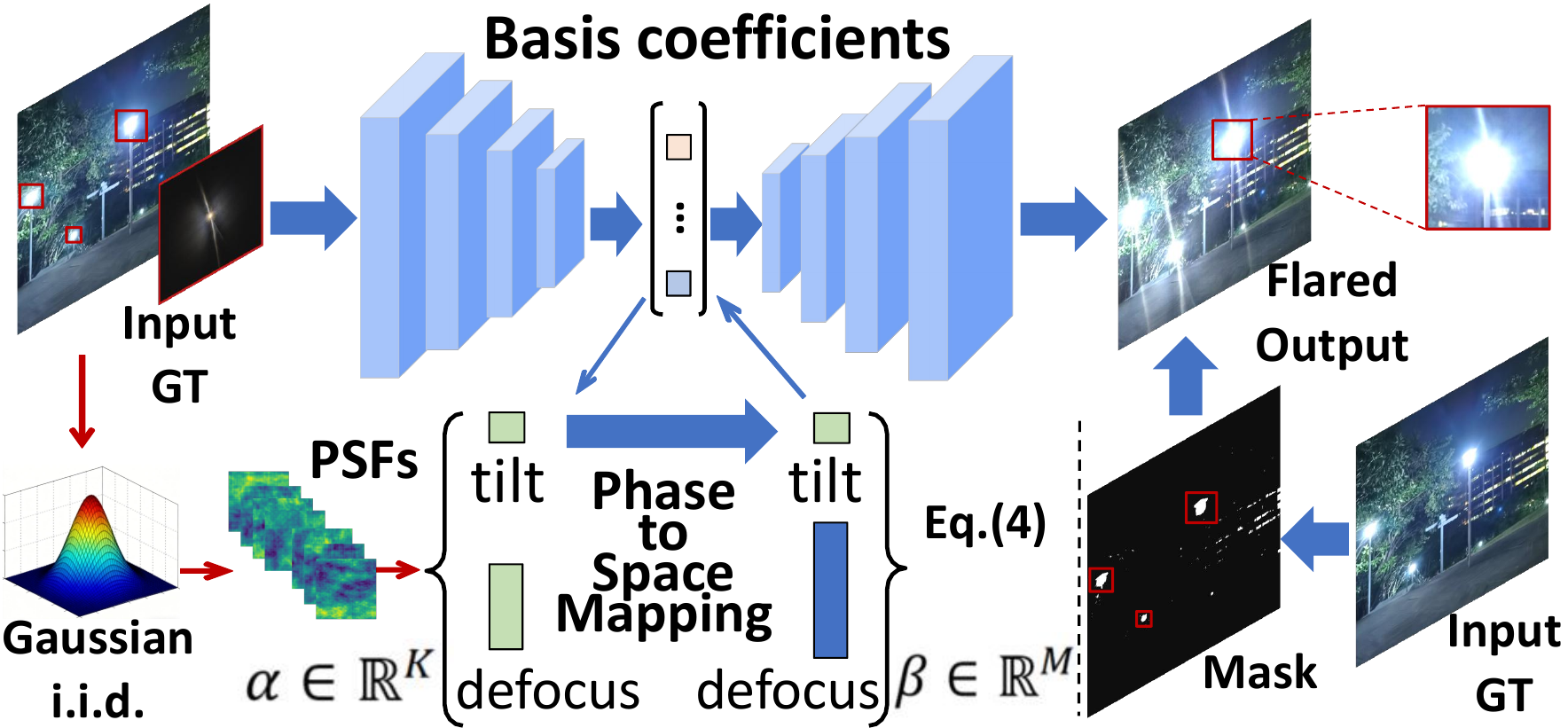}
\caption{The ZernikeVAE-based flare synthesis pipeline. The simulator generates physically realistic flare images by modeling multi-dimensional PSFs with Zernike polynomials and Fourier optics.} 
\label{fig:Zernike}
\end{figure}

Other datasets have explored different aspects of the problem \cite{towardReal, 2024harmonizing, he2025}. \cite{Qiao2021} collected unpaired flare-damaged images and pairs of flare-free images, which enabled an unpaired learning strategy but made it difficult to train pixel-level models. \cite{Deng2024TowardsBlindFlare} proposed the WiderFlare dataset, a real-world dataset customized for flare removal benchmarking in a blind environment, which contains flare severity annotations for each image. However, many existing datasets rely only on gamma-corrected images, ignoring non-linear FSP operations, all of which significantly affect the perceived shape and visibility of flares.

\begin{figure*}[t] 
\centering
\includegraphics[width=\linewidth]{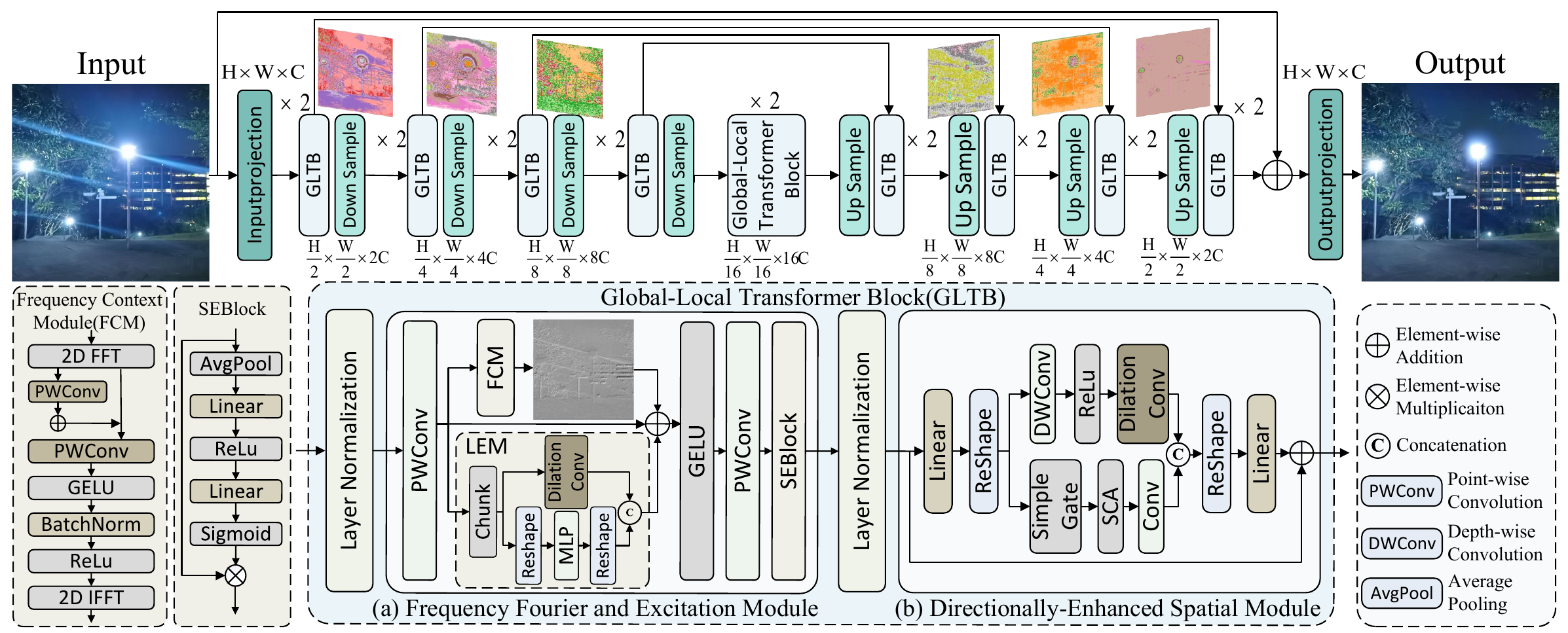}
\caption{Overview of the proposed network architecture. The network adopts a U-shaped encoder-decoder structure, where each stage consists of Global-Local Transformer Blocks (GLTBs). (a) The detailed architecture of Frequency Fourier and Excitation Module FFEM, which integrates frequency representations with channel-wise excitation for efficient global modeling. (b) Directionally-Enhanced Spatial Module (DESM) enhances the feed-forward network by integrating directional and spatial structural information through convolutional design.}  
\label{fig:model}
\end{figure*}

\subsection{Model for Flare Removal}
On the basis of these datasets, various models have been proposed to solve the flare removal problem. Early studies focused on removing masking flares \cite{debevec2004high,seibert1985removal}. These methods, which usually rely on physical priors and deconvolution techniques, work well for smooth artifacts but fail for complex flare patterns.

The advent of deep learning has led to a new wave of methods \cite{2012Alexnet}. \cite{HowToTrain} used their semi-synthetic dataset to train the network with U-Net \cite{Unet} as the backbone. However, its reliance on traditional thresholding makes the model susceptible to unsaturated light sources. For dazzle diversity and model generalization, the latest methods propose end-to-end frameworks. These methods achieve better light source protection and flare suppression \cite{Flare7Kpp}.

Recent methods integrate stronger architectural priors and perceptual constraints \cite{Qiao2021, matta2024gn, zhang2025flare, zhou2025, wanyuwu}. For example, LPFSformer \cite{LPFSformer} introduced a location prior to guide learning, improving positional dependency modeling. FF-Former \cite{FFFormer} combined Swin Transformer and Fourier processing for better global and high-frequency flare removal. MFDNet \cite{Mfdnet} reduced computational cost by multi-band processing instead of direct image correction, while Sparse-UFormer \cite{Sparse-UFormer} used a sparse transformer to capture multi-scale features and focus on flare-relevant regions. However, there are still difficulties in generating more realistic data and handling complex flares, and we improve the synthesis method and design a stronger network module to enhance the flare removal effect.

\section{Dataset}
\subsection{Atmospheric Turbulence Modeling}
Atmospheric turbulence modeling is fundamental for simulating realistic optical distortions in computer vision and computational photography. Specifically, the turbulence-induced phase distortion $\phi(x,y)$ is modeled as a stochastic process with spatial frequency characteristics described by the refractive index structure constant $C_n^2$, leading to random perturbations in the optical wavefront.

Recent simulation frameworks adopt Zernike polynomial decomposition \cite{mao2021turbulence,chimitt2022turbulence} to efficiently approximate these distortions. The resulting point spread function (PSF) is then computed by taking the squared magnitude of the Fourier transform of the complex exponential of the phase. This approach effectively captures diffraction and spatially varying blur characteristics without requiring expensive full-wave simulations.

\subsection{ZernikeVAE-Based Scatter Flare Generation}
Although the Scattering flare in Flare7K dataset enables efficient flare synthesis for supervised training, it lacks physical realism because real-world flares often exhibit spatially varying distortions caused by lens aberrations, diffraction, and micro-scale scattering effects. Consequently, conventional synthetic flares tend to appear overly sharp, lacking the natural smoothness gradients and localized blur complexity of real nighttime scatter flares.

A widely used point spread function (PSF)  $h(\mathbf{x})$  for flare synthesis pipeline is defined with the pupil function at output plane of aperture: 
\begin{align}
    h(\mathbf{x}) &= | \mathcal{F}\left\{ P(\mathbf{x}) \right\} |^{2}\\
P(\mathbf{x}) &= A(\mathbf{x})\exp(j\phi(\mathbf{x}))
\end{align}
where $\mathcal{F}$ denotes the Fourier transform, and $A(\mathbf{x})$ is the aperture function. Since PSFs adopted in Flare 7K \textbf{cite} is mainly used for stimulation of single light resources without interactions, the four types of aperture function $A(\mathbf{x})$ are approximated with PSFs as the sum of their Fourier transformations, which is observed with four separate symmetry PSF patterns. This assumption is not always held, since flare lies between two light sources is affected and exaggerated with both light sources with possible local glow blurring. To better describe this interacted flare phenomena, we revisit the locally flare degradation image $I$ from flare-free input $I_{clear}$ with a set of spatially varying PSFs:
\begin{align}
I\left(\mathbf{x}_i\right) &=\sum_{j=1}^N h_{\mathbf{x}_i}\left(\mathbf{u}_j\right) I_{clear}\left(\mathbf{u}_j\right), \quad i=1, \ldots, N.\\
h_{\mathbf{x}}(\mathbf{u}) &=\left|\mathcal{F}\left(A(\mathbf{x})e^{-j 2 \pi \phi_{\mathbf{x}}(\boldsymbol{\rho})}\right)\right|^2
\end{align}
Here, the phase function $\phi_{\mathbf{x}}(\boldsymbol{\rho})$ is defined per pixel at each coordinate $\mathbf{x}$. The variable $\boldsymbol{\rho} \in \mathbb{R}^2$ is the phase coordinate in the Fourier space.

\begin{equation}
\phi_{\mathbf{x}}(\boldsymbol{\rho}) = \sum_{i=1}^{N} a_i Z_i(\rho)
\label{a}
\end{equation}
where $a_i$ denotes the number of PSF basis. $Z_i(\rho)$ are the orthogonal Zernike polynomials representing various aberration modes, including tilt, defocus, astigmatism, coma, and higher-order distortions. To simplify the equation (\ref{a}) in numerical formation, the PSFs in a spatial decomposition are introduced in cite: 
\begin{equation}
h_{\mathbf{x}}(\mathbf{u})=\sum_{i=1}^N \beta_{ i} \boldsymbol{\varphi}_i(\mathbf{u})
\end{equation}
where $\boldsymbol{\varphi}_i$ spatial basis functions for the PSFs with  locally pixel-based coefficients $\beta_{i}$. This equation enables the local blurring and uneven flare described in the degraded images.

Unlike traditional scalar PSF models, our simulator constructs a multi-dimensional representation, embedding each sampled PSF into a high-dimensional learned kernel space using precomputed basis dictionaries derived from empirical optical data. 

After simulation, our pipeline adopts an encoder-decoder architecture to refine the generated flare. The encoder processes the concatenated Zernike coefficients and kernel size map to estimate the latent distribution's mean $\mu$ and log-variance $\log \sigma^2$. A latent code $z$ is then sampled via the reparameterization trick to model stochastic optical distortions. The decoder reconstructs the final ZernikeVAE flare, improving SLCFormer's generalization for nighttime flare removal.

\begin{figure*}[t] 
\centering
\includegraphics[width=\linewidth]{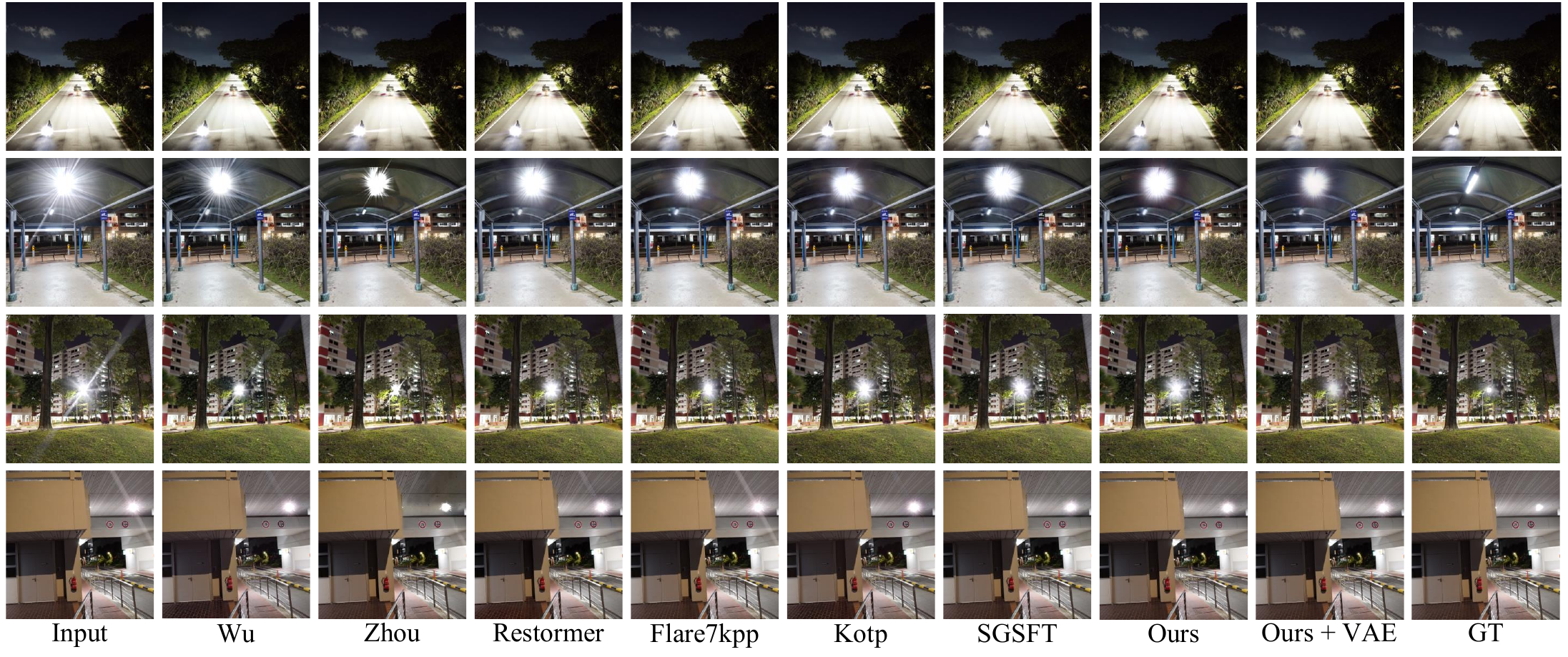}
\caption{Visual comparison of flare removal on synthetic nighttime flare images.} %
\label{fig:synthetic}
\end{figure*}

\begin{figure*}[t] 
\centering
\includegraphics[width=\linewidth]{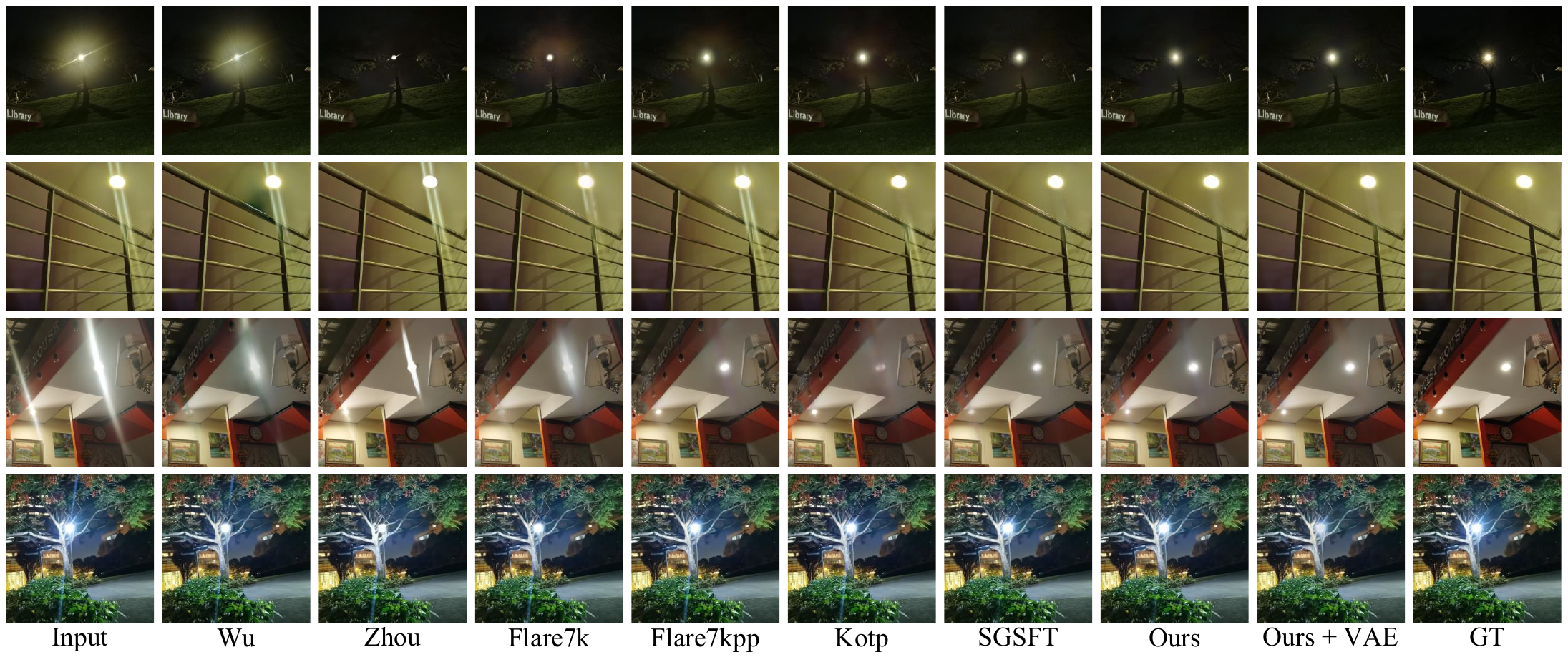}
\caption{Visual comparison of flare removal on real-world nighttime flare images.} %
\label{fig:quality}
\end{figure*}

\section{Flare Removal Framework}
\subsection{Overall Architecture}
As shown in Fig. \ref{fig:model}, our proposed network follows a hierarchical U-shaped encoder-decoder architecture, designed to effectively remove lens flare while preserving structural integrity. Given a corrupted-flare image $I_\text{flare} \in \mathbb{R}^{H \times W \times 3}$, our goal is to reconstruct a clean image $I_\text{deflare} \in \mathbb{R}^{H \times W \times 3}$. To achieve this, we first apply a convolutional feature extractor to obtain an initial feature representation $F_s \in \mathbb{R}^{H \times W \times C}$, which is fed into the encoder path for hierarchical processing.

Each stage of the encoder and decoder is composed of a Global-Local Transformer Block (GLTB), which integrates two key modules: the Frequency Fourier and Excitation Module (FFEM) and the Directionally-Enhanced Spatial Module (DESM). Specifically, FFEM is designed to enhance global context modeling by leveraging spatial and frequency representations. In contrast, DESM complements this by focusing on local structural enhancement and spatially adaptive attention.

To preserve multiscale information and ensure spatial consistency, we adopt pixel unshuffle and pixel shuffle for downsampling and upsampling. Moreover, long-range skip connections are employed to bridge corresponding encoder and decoder stages.

Finally, the decoder reconstructs the output, and a residual connection between the input and output is applied before the final sigmoid activation to produce the deflared image.
 
\subsection{Frequency Fourier and Excitation Module(FFEM)}
FFEM is designed to efficiently capture global contextual representations through frequency-domain analysis, complemented by local enhancement to preserve structural precision. The proposed module comprises three core components: a Local Enhancement Module (LEM), a Frequency Context Module (FCM), and a Squeeze-and-Excitation (SE) Block, which the architecture is shown in Fig. \ref{fig:model}(a).

We first apply a point-wise convolution to the input feature $X \in \mathbb{R}^{H \times W \times C}$, generating an intermediate representation F, which is then split into two parts along the channel dimension:
\begin{align}
F_{\text{local}}, F_{\text{global}} = \text{Split}(\text{PWConv}(X))
\end{align}
the two branches are then processed separately: $F_{\text{local}}$ is fed into LEM to extract spatially localized features, while $F_{\text{global}}$ is passed to FCM to capture long-range dependencies in the frequency domain. The processed features are then concatenated and passed through a GELU activation and another pointwise convolution to fuse them:
\begin{align}
X = \text{PWConv}\Big(\text{GELU}\big(\text{LEM}(F_{\text{local}}) \| \text{FCM}(F_{\text{global}})\big)\Big)
\end{align}
finally, the features $X$ are further refined by a Squeeze-and-Excitation (SE) Block, resulting in the final output. Next, we elaborate on the internal structure and functionality of the three core components: FCM, LEM and the SEBlock.

\begin{table*}[t]
\centering
\setlength{\tabcolsep}{1pt} 
\begin{tabular}{c|ccccc|ccccc}
\toprule
\textbf{Test}& \multicolumn{5}{c|}{\textbf{Real\_Images}} & \multicolumn{5}{c}{\textbf{Synthetic\_Images}} \\
\cmidrule(r){1-1} \cmidrule(lr){2-6} \cmidrule(lr){7-11}
\textbf{Metrics} & PSNR$\uparrow$ & SSIM$\uparrow$ & LPIPS$\downarrow$ & G-PSNR$\uparrow$ & S-PSNR$\uparrow$ & PSNR$\uparrow$ & SSIM$\uparrow$ & LPIPS$\downarrow$ & G-PSNR$\uparrow$ & S-PSNR$\uparrow$ \\
\midrule
Wu~\cite{HowToTrain} & 24.613 & 0.871 & 0.0598 & 21.772 & 16.728 
                     & 28.260 & 0.954 & 0.0331 & 24.757 & 21.108 \\
Zhou~\cite{improvinglens} & 25.149 & 0.883 & 0.0576 & 22.053 & 17.865 
                          & 28.779 & 0.939 & 0.0286 & 23.779 & 22.237  \\
\midrule
U-Net (Ronneberger et al. 2015)   & 27.189 & 0.894 & 0.0452 &                                      23.527 & 22.647 
                                 & 29.389 & 0.933 &  0.0276 &     24.271 & 23.338 \\
HINet~\cite{HINet} & 27.548 & 0.892 & 0.0464 & \textit{24.081} 
                   & \textit{22.907} 
                   & 29.462 & 0.938 & 0.0248 & 24.454 & 23.534 \\
Restormer*~\cite{zamir2022restormer} & 27.597 & \textit{0.897} & 0.0447 
                                     & 23.828 & 22.452 
                        & 29.489 & 0.958 &  0.0244 & 24.654 & 23.078 \\
\midrule
Flare7K~\cite{Flare7k}  & 26.978 & 0.890 & 0.0466 & 23.507 & 21.563 
                        & - & - & - & - & - \\
Flare7Kpp~\cite{Flare7Kpp} & 27.633 & 0.894 & 0.0428 & 23.949 & 22.603 
                       & 29.498 & \textit{0.962} & 0.0210 & 24.685 & 24.155 \\
Flare-Free~\cite{flarefree} & \textit{27.662} & \textit{0.897} & \textit{0.0422} 
                            & \textit{23.987} & \textit{22.847}
                            & \textit{29.573} & 0.961 & \textit{0.0205} & \textbf{24.879} & \textit{24.458} \\
SGSFT~\cite{SGSFT} & \underline{28.077} & \underline{0.904} & \underline{0.0416} & \underline{24.477} & \textbf{23.305}
                            & \underline{29.576} & \underline{0.966} & \underline{0.0200} & \textit{24.745} & \textbf{24.914} \\
\midrule
\textbf{SLCFormer (ours)} & \textbf{28.092} & \textbf{0.905} & \textbf{0.0400} & \textbf{24.497} & \underline{23.287} 
    & \textbf{29.798} & \textbf{0.968} & \textbf{0.0195} & \underline{24.792} & \underline{24.578}\\
\bottomrule
\end{tabular}
\caption{
Quantitative comparison on both the real and synthetic tests from Flare7K++. 
Bold values indicate the best performance among all methods, while underlined values indicate the second-best and italicised values indicate the third-best.
Models marked with "*" have reduced parameters due to GPU memory limitations. 
Note that since all models are trained on Flare7K++, and the architecture of Flare7K is exactly the same as Uformer but trained on Flare7K instead of Flare7K++, the results of Flare7K on the synthetic test set are omitted.
}
\label{tab:quantitaive}
\end{table*}

\subsubsection{Frequency Context Module (FCM).}
The Fourier Transform provides an efficient way to capture global structures by converting spatial features into frequency components. The Fast Fourier Transform (FFT) enables holistic frequency-domain analysis with significantly lower computational cost ($\mathcal{O}(N \log N)$) than spatial-domain methods such as multi-head self-attention (MSA)($\mathcal{O}(N^2)$). MSA suffers from redundant computations in modeling global relationships. In contrast, FFT compactly represents global structures through orthogonal frequency bases, avoiding redundancy while maintaining scalability. Thus, we adopt FFT as an efficient substitute for MSA.

The FCM first performs the 2D Fast Fourier Transform to project the input $X \in \mathbb{R}^{H \times W \times C}$ into the frequency domain. After frequency transformation, the real and imaginary parts are concatenated along the channel dimension and processed by two lightweight 1×1 convolutions and a non-linear activation. The final frequency representation is then projected back to the spatial domain using the inverse FFT. The entire transformation process can be summarized as follows:
\begin{align}
F &= \mathcal{F}(X) \\
F &= \text{Conv}_{1 \times 1}(F) + F \\
F &= \text{GELU}(\text{Conv}_{1 \times 1}(F)) \\
X &= X + \mathcal{F}^{-1}\bigl(\text{BN}(\text{ReLU}(F))\bigr)
\end{align}
where $\mathcal{F}(\cdot)$ and $\mathcal{F}^{-1}(\cdot)$ denote a fast Fourier transform (FFT) and an inverse transform (IFFT), respectively; $\text{Conv}_{1 \times 1}(\cdot)$ denotes a 1×1 convolution operation in the frequency domain; GELU and ReLU are nonlinear activation functions; and BN denotes a batch normalization operation. In addition, the final output is added back to the original input through residual concatenation X to enhance global modeling while preserving structure information.

\subsubsection{Local Enhancement Module (LEM).}
While FFT excels at capturing long-range dependencies, its global receptive field tends to overlook certain local details. To address this limitation, we introduce LEM to compensate for local feature extraction. LEM operates on the input feature $F_{\text{local}} \in \mathbb{R}^{B \times C \times H \times W}$, and first splits it equally into two parts: $F_1 \in \mathbb{R}^{B \times \frac{C}{2} \times H \times W}$ and $F_2 \in \mathbb{R}^{B \times \frac{C}{2} \times H \times W}$. The first half is passed through a spatially independent token-wise MLP. $F_2$ is processed using a dilated convolution to capture spatial dependencies. The two features are then concatenated along the channel axis to form the locally enhanced output.

\subsubsection{Squeeze-and-Excitation (SE) Block.}
To improve network representation, we integrate a Squeeze-and-Excitation (SE) \cite{SEBlock} block after the module.

Given an input feature $F \in \mathbb{R}^{B \times H \times W \times C}$, the SE block first applies a global average pooling to generate a channel descriptor. This descriptor passes through two fully connected layers with ReLU and sigmoid activations, obtaining channel weights $S \in \mathbb{R}^{B\times C}$, which are then multiplied $F$ to produce the refined output. The entire process can be summarized as follows:
\begin{align}
    F = F \odot \sigma(w_2(\delta(w_1(\mathrm{AvgPool}(F)))))
\end{align}
here, $\delta$ and $\sigma$ are ReLU and sigmoid, and $\odot$ denotes channel-wise multiplication.

\subsection{Directionally-Enhanced Spatial Module (DESM)}
Traditional FFNs often lack the ability to capture spatial structure and directional cues, which are essential for accurately restoring flare-degraded regions. DESM aims to introduce spatially-aware interactions within the FFN by injecting directional and local structural information through convolutional design. 

As shown in Fig. \ref{fig:model}(b), input tokens are projected to a higher-dimensional space and reshaped into 2D feature maps, then split into two branches. One applies depth-wise and dilated convolutions ($X_{\text{dir}}$) to encode spatial and directional features, while the other uses a SimpleGate with channel attention ($X_{\text{gate}}$) to enhance structural representations. The outputs are concatenated and linearly projected:
\begin{align}
X = \text{Linear}(\text{Concat}(X_{\text{dir}}, X_{\text{gate}}))
\end{align}

\subsection{Loss Function}
In order to efficiently supervise the flare removal, we employ an integrated loss function with the following expression:

\begin{align}
    L = \lambda_1 L_1 + \lambda_2 L_{\text{vgg}} + \lambda_3 L_{\text{rec}} + \lambda_4 L_{\text{hf}}
\end{align}
where $L_1$, $L_{\text{vgg}}$, $L_{\text{rec}}$ denote the standard $L1$ loss, the perceptual loss \cite{simonyan2014VGG}, and the reconstruction loss \cite{Flare7Kpp}, respectively.

To enhance the recovery of high-frequency details, we further introduce a high-frequency loss $L_{\text{hf}}$ combining Laplacian and Sobel gradients to capture fine flare structures:
\begin{align}
L_{\text{hf}} = \tfrac{1}{2}\|\nabla_{\!\text{lap}}(\hat{I}){-}\nabla_{\!\text{lap}}(I)\| + \tfrac{1}{2}\|\nabla_{\!\text{sob}}(\hat{I}){-}\nabla_{\!\text{sob}}(I)\|
\end{align}
where $\hat{I}$ and $I$ are the predicted and ground-truth images, and $\nabla_{\text{lap}}, \nabla_{\text{sob}}$ denote Laplacian and Sobel operators. The weighting coefficients, $\lambda_1$, $\lambda_2$, $\lambda_3$, $\lambda_4$ are empirically set to 0.5, 0.5, 1.0, and 1.0, respectively.

\begin{figure}[t]
    \centering
    \includegraphics[width=\linewidth]{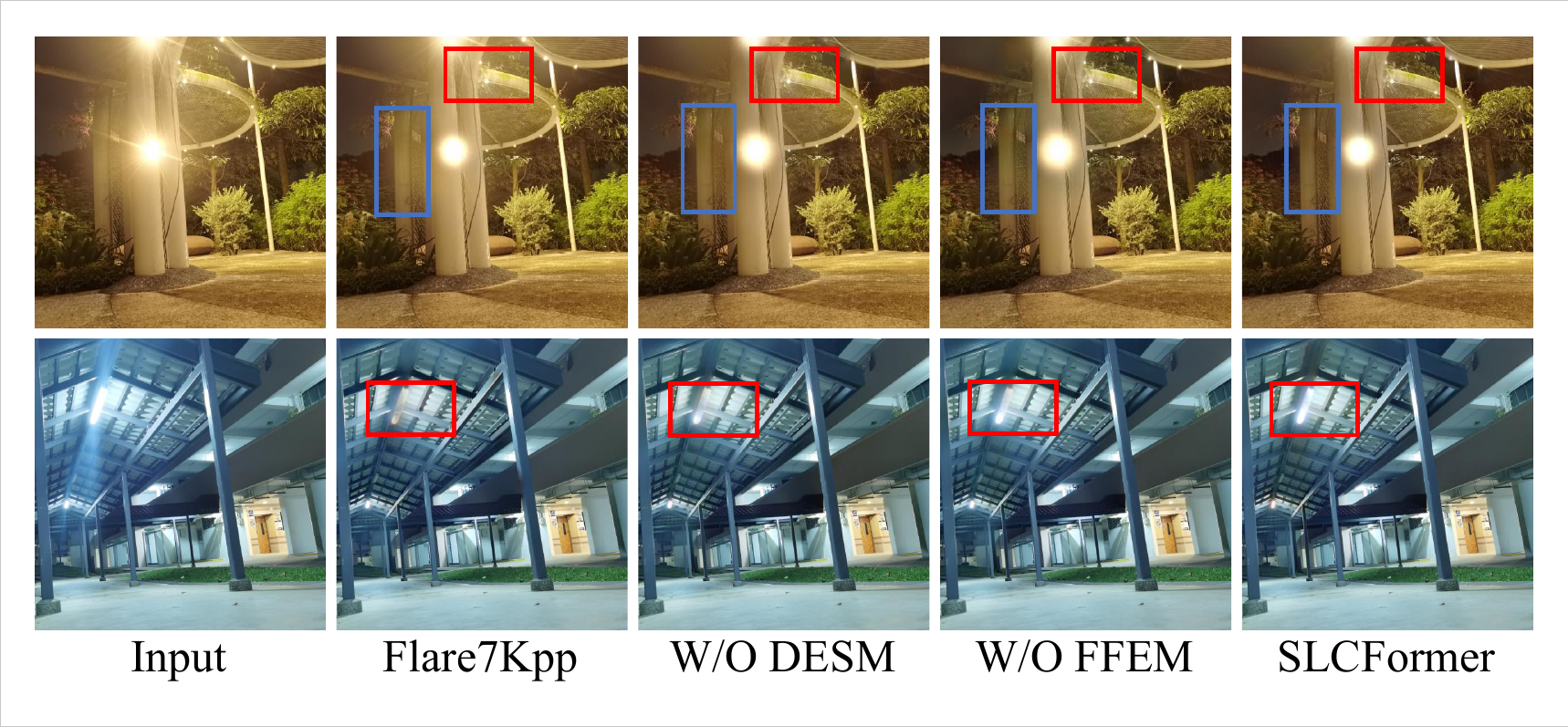} 
    \caption{Ablation studies on the proposed method.}
    \label{fig:ablation1} %
\end{figure}
\begin{figure}[t]
    \centering
    \includegraphics[width=\linewidth]{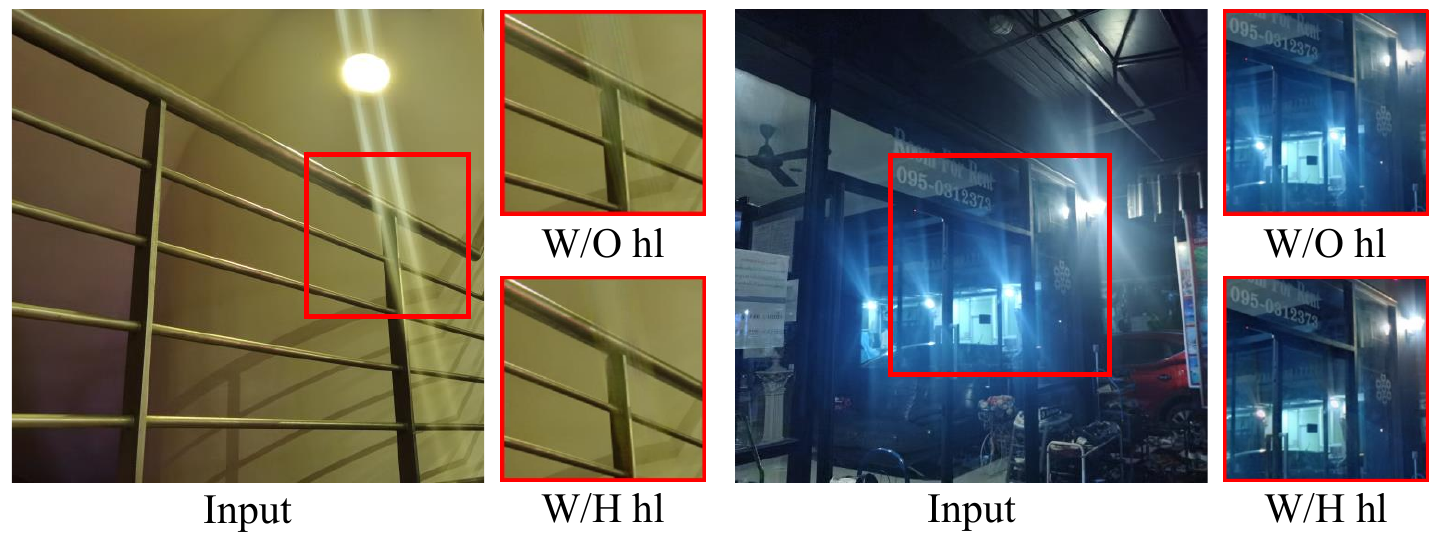} 
    \caption{Ablation studies on the proposed loss function.}
    \label{fig:ablation2} %
\end{figure}
\begin{table}[t]
\centering
\setlength{\tabcolsep}{1.5pt} 
\begin{tabular}{l|ccccc}
\toprule
\textbf{Method} & PSNR$\uparrow$ & SSIM$\uparrow$ & LPIPS$\downarrow$ & G-PSNR$\uparrow$ & S-PSNR$\uparrow$ \\
\midrule
W/O Both   & 27.633 & 0.894  & 0.0428 & 23.949 & 22.603 \\
W/O FFEM   & 27.888 & 0.902  & 0.0408 & 24.473 & 23.137 \\
W/O DESM   & 27.804 & 0.899  & 0.0430 & 24.372 & 22.926 \\
Full Model &\textbf{28.092} & \textbf{0.905}  & \textbf{0.0400} & \textbf{24.497} & \textbf{23.287} \\

\bottomrule
\end{tabular}
\caption{Ablation studies on the proposed method.}
\label{tab:ablation1}
\end{table}
\begin{table}[t]
\centering
\setlength{\tabcolsep}{2.5pt} 
\begin{tabular}{l|ccccc}
\toprule
\textbf{Method} & PSNR$\uparrow$ & SSIM$\uparrow$ & LPIPS$\downarrow$ & G-PSNR$\uparrow$ & S-PSNR$\uparrow$ \\
\midrule
W/O $L_{\text{hf}}$ & 27.824 & 0.898  & 0.0431 & 24.302 & 22.986 \\
W/H $L_{\text{hf}}$ & \textbf{28.092} & \textbf{0.905}  & \textbf{0.0400} & \textbf{24.497} & \textbf{23.287} \\
\bottomrule
\end{tabular}
\caption{Ablation study on the proposed loss function.}
\label{tab:ablation2}
\end{table}

\section{Experiments}
\subsection{Datasets}
To train our flare removal model, we construct a supervised training set based on the Flare7K++ dataset \cite{Flare7Kpp}, excluding 5,000 scattering flares from Flare7K and 964 from Flare-R. Background images are sampled from Flickr-24K \cite{zhang2018Flikr24K}. To ensure that the network learns to restore both the underlying scene and the light source, we use the light source.

To enhance realism and diversity, we apply a series of photometric and geometric augmentations to flare and background images during synthesis. These operations improve the robustness of the model by simulating a wide range of flare appearances. Detailed augmentation parameters are provided in the supplementary material.

\subsection{Implementation Details}
We implement our model using the PyTorch framework and train it on an NVIDIA RTX 3090 GPU. During training, both flare-corrupted and flare-free input images are cropped to a fixed resolution of 512×512. We use a batch size of 2 and adopt the Adam optimizer with $\beta_1 = 0.9$, $\beta_2 = 0.99$. The initial learning rate is set to 1e-4 and is scheduled using a MultiStepLR strategy, with the learning rate reduced by a factor of 0.5 after 200K iterations. The total number of training iterations is 400K.

\subsection{Evaluation Metrics}
To comprehensively evaluate the performance of our model, we adopt several restoration metrics, including PSNR, SSIM\cite{wang2004SSIM}, and LPIPS\cite{zhang2018LPIPS}. In addition, we incorporate two specialized metrics proposed by \cite{Flare7Kpp}, namely S-PSNR and G-PSNR, which are designed to assess the effectiveness of flare removal in localized streak and glare regions, respectively.

\subsection{Comparison with Previous Methods}
To validate the advantages of our proposed SLCFormer, we conduct both qualitative and quantitative evaluations against a range of 
flare removal and image restoration methods on both the Flare7K real and synthetic test dataset, including U-Net \cite{Unet}, HINet \cite{HINet}, Restormer \cite{zamir2022restormer}, Uformer \cite{uformer}, etc.

\subsubsection{Qualitative Evaluation.}
We compare visual results on both real-world nighttime flare and synthetic flare removal in Fig. \ref{fig:synthetic} and Fig. \ref{fig:quality}. The recovered images produced by SLCFormer exhibit noticeably cleaner flare removal. Moreover, SLCFormer trained with VAE-augmented scatter flares produces more realistic outputs.

\subsubsection{Quantitative Evaluation.}
Table \ref{tab:quantitaive} summarizes the full-reference metrics on the Flare7K real and synthetic tests. SLCFormer achieves a PSNR of 28.092 dB in the real test images. In SSIM, our model reaches 0.905. For perceptual fidelity, SLCFormer attains the lowest LPIPS score of 0.0400, indicating closer alignment with ground-truth perceptual quality.

\subsection{Ablation Study}
\subsubsection{Network Structure.}
We further illustrate qualitative comparisons in Figure \ref{fig:ablation1}. In the first group of images, results produced by the full model appear cleaner and less hazy, while other configurations show varying degrees of turbidity. Although w/o FFEM generates relatively clearer results compared to other ablated models, it still fails to match the clarity of the full model. Additionally, the shimmer in the scene is effectively removed only by the full model and w/o DESM, whereas other models fail to eliminate it.

Additionally, only the full model preserves the light source region completely intact in the second group. Other models remove parts of the light source, leading to unnatural results. These results demonstrate that both modules are essential for flare removal.

\subsubsection{Loss Function.}
We further evaluate the impact of the proposed high-frequency loss by conducting ablation experiments. As shown in Table \ref{tab:ablation2}, incorporating the high-frequency loss leads to consistent improvements across all evaluation metrics. Qualitative results in Figure \ref{fig:ablation2} show that incorporating the high-frequency loss yields sharper details and clearer structures, with shimmer flares more effectively suppressed and overall flare intensity visibly reduced.

\section{Conclusion}
In this paper, we propose a ZernikeVAE-based scatter flare generation pipeline that synthesizes physically realistic flare patterns by modeling spatially varying point spread functions (PSFs) with interpretable Zernike polynomials. This approach enriches the diversity and complexity of flare distributions, thereby improving data realism and enhancing the generalization capability of learning-based flare removal models. Based on this, we introduced SLCFormer, integrating the Frequency Fourier and Excitation Module (FFEM) and Directionally-Enhanced Spatial Module (DESM) to jointly model global frequency context and local structural features for effective nighttime flare removal. Experiments demonstrate that our model outperforms most current methods, though it introduces additional computational overhead and the ZernikeVAE-based synthesis may not fully capture real flare complexity. Future work will explore lightweight architectures and more realistic flare synthesis to extend applicability to broader optical degradation restoration tasks.

\section{Acknowledgments}
This work was supported financially by the Natural Science Foundation of China (62202347).

\bibliography{aaai2026}

\clearpage 
\appendix  
\twocolumn[
    \centering
    \LARGE \textbf{Appendix} 
    \vspace{5em}             
]
\section{Model Details}
\subsection{Datasets}
To ensure that the synthesized flare training data exhibits sufficient realism, diversity, and robustness, we apply a set of controlled photometric and geometric transformations to both the flare and background images before composition.The specific augmentation parameters used in our pipeline are listed below(Table \ref{tab:data_aug}):
\begin{table}[H]
\centering
\setlength{\tabcolsep}{1.5pt}
\begin{tabular}{l|l|l}
\toprule
\textbf{Augmentation Type} & \textbf{Parameter Distribution} & \textbf{Applied To} \\
\midrule
 Inverse Gamma           & $\gamma \sim U(1.8,\ 2.2)$ & Both \\
RGB Scaling              & $U(0.5,\ 1.2)$             & Both \\
Gaussian Noise           & $\sigma^2 \sim 0.01\chi^2$ & Background \\
Intensity Offset         & $U(-0.02,\ 0.02)$          & Flare \\
Color Jitter             & $U(0.8,\ 3.0)$             & Flare \\
\midrule
Rotation                 & $\theta \sim U(0,\ 2\pi)$  & Flare \\
Translation              & $U(-300,\ 300)$            & Flare \\
Shear                    & $U(-20^\circ,\ 20^\circ)$  & Flare \\
Scaling                  & $U(0.8,\ 1.5)$             & Flare \\
\midrule
Gaussian Blur            & $\sigma \sim U(0.1,\ 3.0)$ & Flare \\
\bottomrule
\end{tabular}
\caption{Data augmentation settings applied to flare and background images during training.}
\label{tab:data_aug}
\end{table}

\subsection{Directionally-Enhanced Spatial Module (DESM)}
We will further elaborate on the design of the $X_{\text{gate}}$ branching below. We first split the map $X_{\text{gate}}$ into two halves along the channel dimension, denoted as $X_1$ and $X_2$. The SimpleGate mechanism performs element-wise gating to selectively enhance informative features:
\begin{align}
X_{\text{sg}} = X_1 \odot X_2
\end{align}
where $\odot$ denotes element-wise multiplication. This operation reinforces structural details.

The purpose of the Spatial Channel Attention (SCA) module is emphasizing important regions. It first applies global average pooling followed by a convolution to generate attention weights:
\begin{align}
S = \text{Conv}(\text{AvgPool}(X_{\text{sg}}))
\end{align}
And the final output of this branch is obtained by rescaling the gated features:
\begin{align}
X_{\text{gate}} = S \odot X_{\text{sg}}
\end{align}

\section{More Experiments}
\subsection{Qualitative Evaluation}
We further conduct qualitative analysis on the flare-corrupted test provided by the Flare7Kpp, which consists of flare-contaminated images without ground truth. This evaluation aims to assess the model's capability in practical, uncontrolled nighttime scenes where flares exhibit highly complex and varying intensities.

\begin{figure*}[t]
    \centering
    \includegraphics[width=\linewidth]{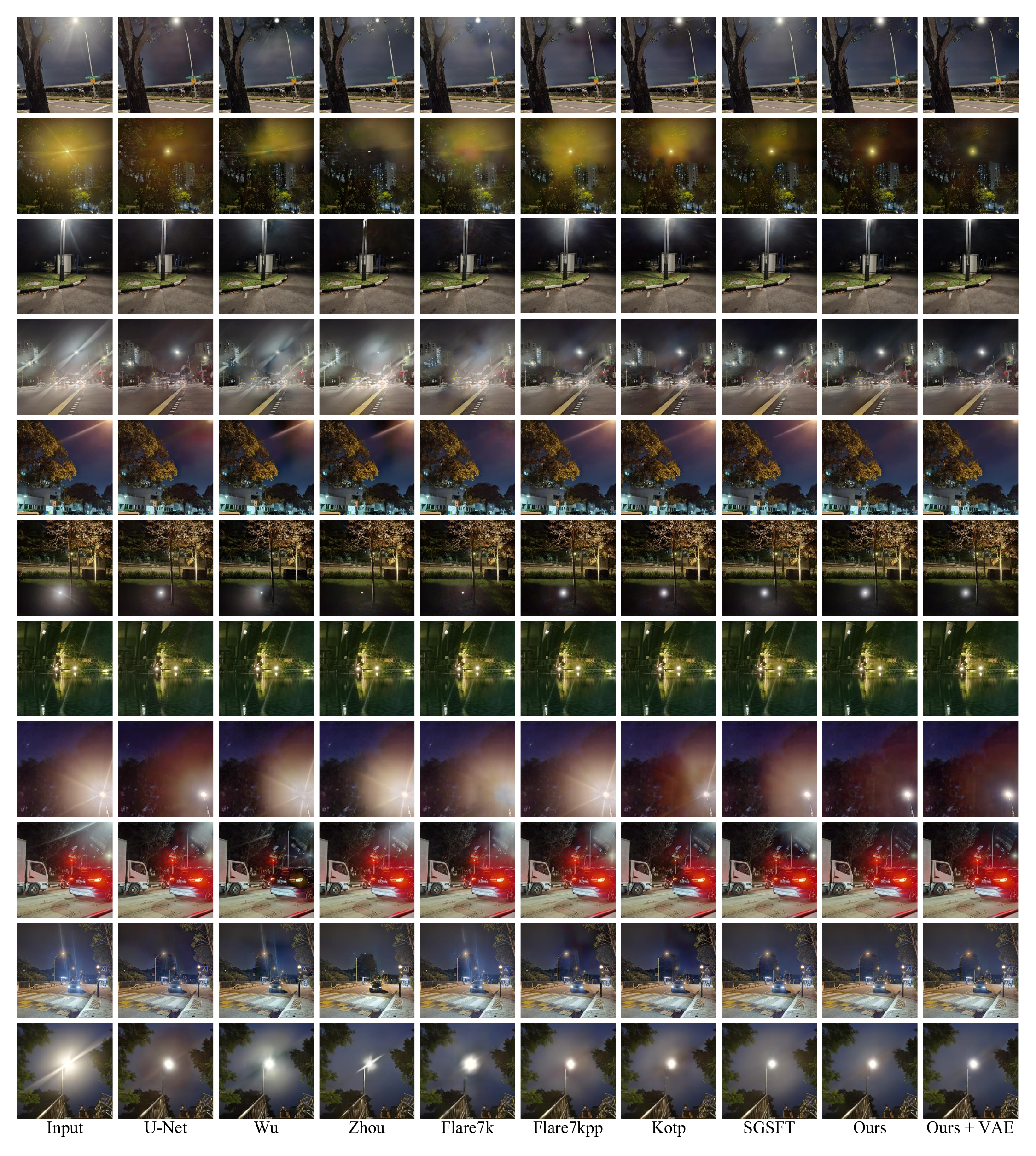} 
    \caption{Visual comparison of flare removal on flare-corrupted test.}
    \label{fig:add1} %
\end{figure*}

As illustrated in Fig. \ref{fig:add1}, compared with previous models, our method produces cleaner and more natural results. It better suppresses directional streaks and scattered light while preserving image details. In challenging flare cases, our model recovers clearer structures with fewer artifacts.

\begin{table}[t]
\centering
\setlength{\tabcolsep}{1.5pt} 
\begin{tabular}{l|ccccc}
\toprule
\textbf{Method} & PSNR$\uparrow$ & SSIM$\uparrow$ & LPIPS$\downarrow$ & G-PSNR$\uparrow$ & S-PSNR$\uparrow$ \\
\midrule
W/O Both   & 29.498 & 0.962  & 0.0210 & 24.685 & 24.155 \\
W/O FFEM   & 29.388 & 0.965  & 0.0209 & 24.478 & 24.443 \\
W/O DESM   & 29.702 & 0.963  & 0.0225 & 24.640 & 24.284 \\
Full Model &\textbf{29.798} & \textbf{0.968} & \textbf{0.0195} & \textbf{24.792} & \textbf{24.578}\\
\bottomrule
\end{tabular}
\caption{Ablation study on the synthetic images.}
\label{tab:add1}
\end{table}

\begin{figure*}[t]
    \centering
    \includegraphics[width=\linewidth]{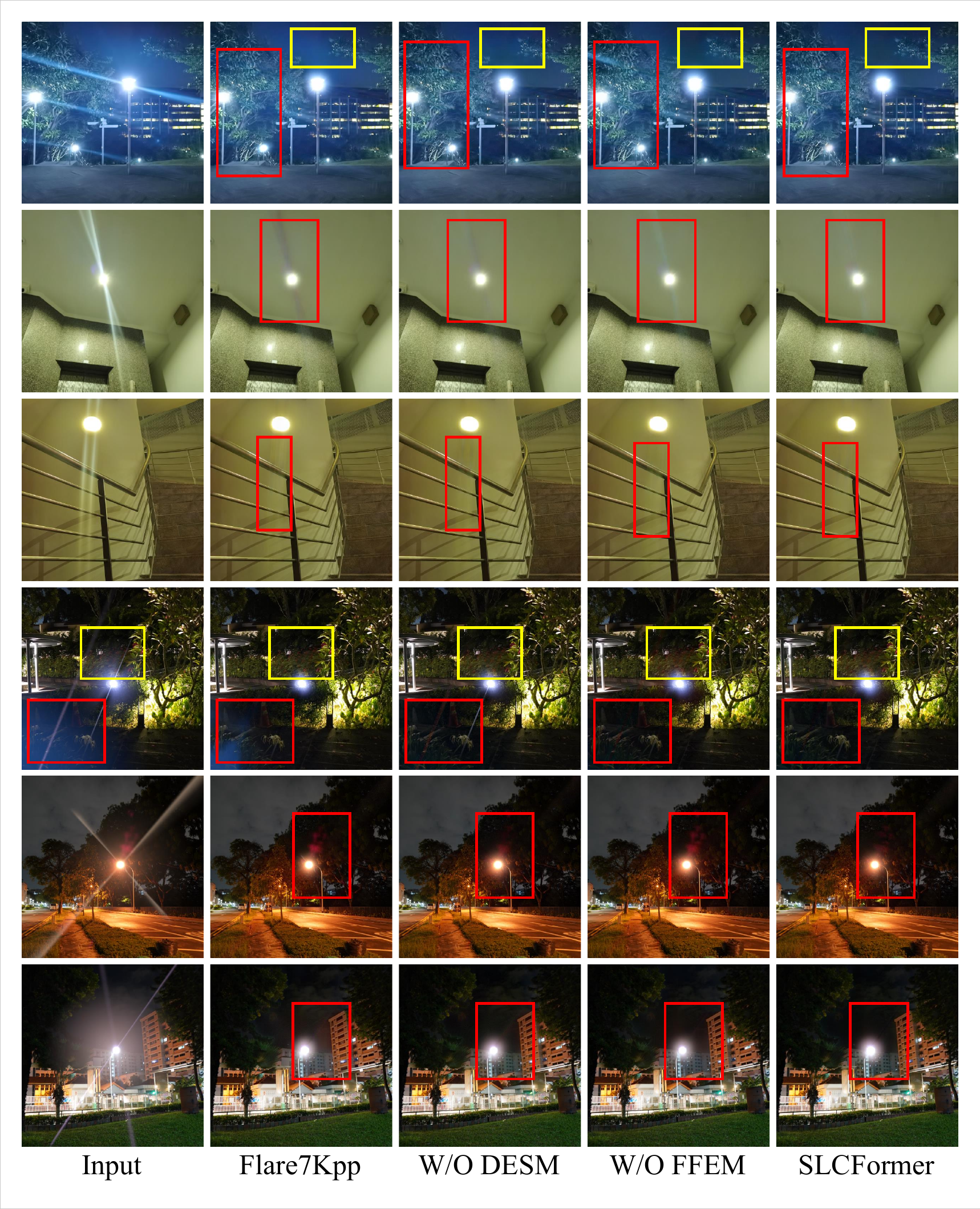} 
    \caption{Additions to the ablation study on the proposed method.}
    \label{fig:add2} %
\end{figure*}

\begin{figure*}[t]
    \centering
    \includegraphics[width=\linewidth]{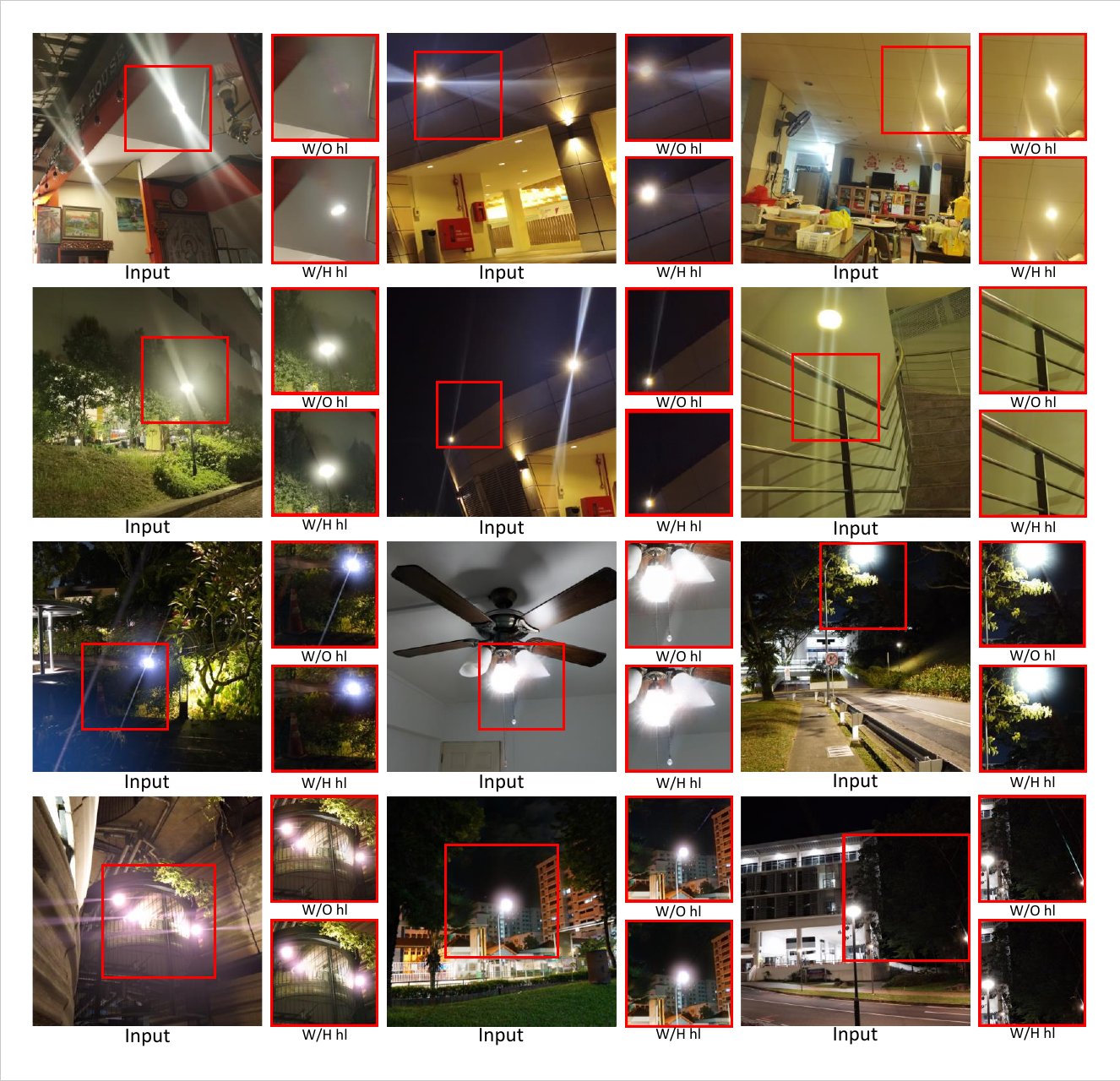} 
    \caption{Additions to the ablation study on the proposed loss function.}
    \label{fig:add3} %
\end{figure*}

\subsection{Ablation Study}
To further validate the effectiveness of our design, we provide additional qualitative results for ablation studies. 

As shown in Fig. \ref{fig:add2}, models equipped with our proposed module produce visually superior results, effectively reconstructing flare-degraded regions. Regarding the loss design, Fig. \ref{fig:add3} shows that incorporating high-frequency loss better removes flare streaks while preserving structural fidelity. These comparisons confirm that both our architectural design and loss functions contribute significantly to improved flare removal.

In addition, we report quantitative results for the module ablation in Table \ref{tab:add1}, using synthetic test. Compared to the baseline, our design brings a slight but consistent improvement in PSNR and SSIM.

\end{document}